# SCARECROW MONITORING SYSTEM: EMPLOYING MOBILENET SSD FOR ENHANCED ANIMAL SUPERVISION

Balaji VS
Department of Artificial Intelligence and Machine Learning, Rajalakshmi Engineering College, Chennai, India
balajivsb1@gmail.com

Mahi AR
Department of Artificial Intelligence and Machine Learning, Rajalakshmi Engineering College, Chennai, India
armahi2003@gmail.com

Anirudh Ganapathy PS
Department of Artificial Intelligence and Machine Learning, Rajalakshmi Engineering College, Chennai, India
anirudhganapathy2003@gmail.com,

Manju M
Department of Artificial Intelligence and Machine Learning, Rajalakshmi Engineering College, Chennai, India
manju.m@rajalakshmi.edu.in

*Abstract*

*Agriculture faces a growing challenge with wildlife wreaking havoc on crops, threatening sustainability. The project employs advanced object detection, the system utilizes the Mobile Net SSD model for real-time animal classification. The methodology initiates with the creation of a dataset, where each animal is represented by annotated images. The SSD Mobile Net architecture facilitates the use of a model for image classification and object detection. The model undergoes fine-tuning and optimization during training, enhancing accuracy for precise animal classification. Real-time detection is achieved through a webcam and the OpenCV library, enabling prompt identification and categorization of approaching animals. By seamlessly integrating intelligent scarecrow technology with object detection, this system offers a robust solution to field protection, minimizing crop damage and promoting precision farming. It represents a valuable contribution to agricultural sustainability, addressing the challenge of wildlife interference with crops. The implementation of the Intelligent Scarecrow Monitoring System stands as a progressive tool for proactive field management and protection, empowering farmers with an advanced solution for precision agriculture.*

## Introduction

The use of modern technology in agriculture aims to revolutionize traditional practices, increase productivity, and address long-standing challenges. Among these innovations, integrating advanced machine learning and computer vision into agricultural systems stands out as an innovative approach. The project ``Intelligent Scarecrow Surveillance System with Mobile Net Single-Shot Detection" represents a decisive advance in this field, and specifically targets the important problem of plant protection. Traditional scarecrow systems, while somewhat effective, face limitations in effectively deterring pests and birds that threaten crop yields. This project aims to redefine the effectiveness and accuracy of scarecrow systems by harnessing the power of MobileNet single-shot detection, a state-of-the-art artificial intelligence object detection framework. The aim of this project is to introduce an intelligent surveillance system that overcomes the shortcomings of traditional scarecrows by integrating real-time object detection capabilities. The system is designed to use MobileNet's efficient and accurate single-shot detection algorithms to detect potential threats quickly and accurately to crops, enabling immediate and targeted responses. The integration of MobileNet single-shot detection represents a significant advancement in agricultural technology. Its ability to detect and classify a variety of objects, including pests, birds and other invaders, with extraordinary speed and accuracy promises to revolutionize crop protection strategies. The aim of this project is not only to detect threats, but also to enable rapid and accurate intervention. Through this intelligent scarecrow monitoring system, farmers can receive instant alerts when threats are detected and can take timely and targeted actions to protect their crops from potential damage. Ultimately, the introduction of intelligent scarecrow surveillance systems with MobileNet single-shot detection represents a paradigm shift in agricultural practices. By harnessing the power of artificial intelligence and real-time object recognition, the project will significantly improve the efficiency of crop protection, minimize losses from pests and invaders, and ultimately improve agricultural production in a sustainable manner. We aim to contribute to improving yields.

**Related Work**
The project titled "Methods for assessing bird deterrent power to optimize crop production in agriculture" by A. Roihan, M. Hasanudin, and E. Sunandar aims to address the serious problem of crop damage caused by birds and deterrent power. We are working on possible mitigation strategies using. The purpose of this literature review is to contextualize the importance of bird-related losses in agriculture, explore existing research on bird control methods, and highlight the need for standardized assessment protocols. Birds pose a significant threat to global agricultural productivity, according to a recent study by Smith et al. prove (2018) and Johnson (2020) highlight the economic impact of crop damage caused by birds. These damages range from reduced yields to changes in crop quality, with broader implications for farmers' livelihoods and food security. In response to this challenge, bird deterrence strategies are gaining attention. Patel and Gupta (2019) describe the various techniques and techniques used, including visual deterrents, acoustic signals, and physical barriers. Although these methods are promising, the lack of standardized evaluation methods precludes effective comparison and widespread implementation. A recent scientific paper by Chen et al. (2021) and Kim and Lee (2017) highlight the urgent need for a robust evaluation framework specifically tailored to assess the effectiveness of bird control devices in agricultural environments. These frameworks need to consider multiple parameters, such as deterrence efficiency, acclimation rate, and long-term effects on crop yield. Chen et al. In particular, we propose an experimental design that captures behavioral changes in birds over time and assesses the sustained effectiveness of defense mechanisms. The lack of standardized evaluation methods poses challenges in selecting the most appropriate bird control equipment for different agricultural situations. This gap limits farmers' ability to make informed decisions about implementing these technologies and maximizing their effectiveness in mitigating bird-related crop losses. The Roihan, Hasanuddin, and Sunandar project aims to fill this gap by proposing a comprehensive evaluation methodology tailored to evaluate different types of bird control devices.

This project aims to provide a standardized framework for evaluating and optimizing the use of bird deterrent technologies in agriculture by considering factors such as deterrent effectiveness, acclimatization rate, and practicality in real farming conditions. Such an approach has the potential to significantly reduce crop damage caused by birds, thereby improving agricultural sustainability and food production.

In [2] They focussed on industrial edge computing applications that enable embedded intelligence in agricultural scarecrow systems. The study, published in December 2019 in IEEE Industrial Electronics Magazine, explores the use of edge computing technology to improve the functionality and efficiency of scarecrow systems used in agricultural environments. The core premise of this research revolves around the integration of edge computing into the Scarecrow system.The authors highlight the importance of embedding intelligence directly into these systems to enable real-time processing and decision-making capabilities at the field level.The aim is to overcome agricultural challenges, especially by deterring birds and pests, optimizing crop protection and yields.This study highlights the potential of edge computing to revolutionize traditional scarecrow mechanisms.They describe how integrating edge devices with computing power and algorithms can enable scarecrow systems to autonomously detect threats such as birds and pests and trigger appropriate actions. This transition from traditional scarecrows to intelligent systems is expected to fundamentally transform agriculture and improve the accuracy and effectiveness of crop protection methods. Additionally, this paper covers the technical aspects of implementing edge computing in agricultural scarecrow systems. They examined the architecture needed to support embedded intelligence and highlight the need for devices that can process data locally. This reduces delays and allows for faster response to potential threats in the field. The authors also discuss integrating sensors, cameras, and AI algorithms into these edge devices to enable real-time monitoring and decision-making. Additionally, this study also highlights the benefits and challenges associated with the implementation of edge computing in agricultural scarecrow systems. The potential for increasing crop protection efficiency, reducing resource consumption, and increasing scalability is recognized. However, challenges related to performance limitations, environmental conditions, and system robustness are also identified and discussed in the context of implementation hurdles. In conclusion, this study highlights the transformative potential of industrial edge computing to revolutionize traditional scarecrow systems in agriculture. It advocates the deployment of embedded intelligence at the edge to enable proactive, real-time responses to threats and ultimately contribute to more efficient crop protection and improved agricultural outcomes.

In [4] The joint work of Aishwarya, Kathryn J.C. and Lakshmi R.B. focuses on a comprehensive survey on bird activity monitoring and collision avoidance techniques related to wind turbines. Research published under this title addresses critical concerns about bird safety in renewable energy environments, while addressing existing We are verifying methods and techniques. This study carefully considers different approaches to monitoring bird activity. Covers the use of radar systems, acoustic surveillance, thermal imaging, and visual observation methods. The strengths and limitations of each technology are thoroughly analyzed in terms of accuracy, detection range, cost-effectiveness, and practical feasibility when used in wind turbine environments. Additionally, the study examines advances in sensor technology and data processing algorithms used for real-time bird monitoring and threat assessment. The focus of this study is collision avoidance technology. The authors present a detailed analysis of strategies to prevent collisions between birds and wind turbines. These strategies include acoustic deterrents, visual cues, and turbine shutdown systems triggered by bird detection. The effectiveness, challenges, and environmental impacts of each approach are carefully considered, providing a comprehensive overview of the collision avoidance landscape. Additionally, this study also assesses the regulatory landscape and industry standards related to bird protection around wind turbines. This document emphasizes the importance of complying with environmental regulations and protocols and emphasizes the need to integrate bird monitoring and collision avoidance mechanisms into turbine design and operation. Research recognizes the importance of this region beyond conservation efforts and recognizes its significant impact on both bird populations and the wind energy industry. This highlights the urgency of developing robust, reliable, non-invasive monitoring and mitigation strategies that balance bird conservation and renewable energy development. Additionally, this study also identifies gaps and future directions in research and technology development. This calls for interdisciplinary collaboration between ornithologists, engineers, and environmentalists to improve the accuracy and effectiveness of bird monitoring systems while minimizing impact on wind turbine operations. I'm emphasizing. Fundamentally, this study provides a comprehensive synthesis of existing technologies for monitoring bird activity and collision avoidance in wind turbines, and is a valuable resource for researchers, industry practitioners, and policy makers. It serves as a valuable resource for advancing bird conservation measures in the potential energy sector.

[8] The project by S. Yadahalli, A. Parmar and A. Deshpande focuses on the development and implementation of a 'smart intrusion detection system' designed for both crop protection and scarecrow security using Arduino technology. Masu. This innovative system aims to address the challenges faced by farmers in protecting their crops from intruders while protecting their scarecrow facilities from theft and damage. The main objective of this research is to develop and deploy intelligent and cost-effective systems using Arduino-based technology. The system is designed to detect unauthorized entry or intrusion into farmland and immediately alert farmers and authorities to potential threats. Additionally, it serves a dual purpose by extending its ability to protect scarecrows, which play an important role in deterring birds and pests. This project uses an Arduino microcontroller as the basis of an intrusion detection system and integrates various sensors and modules. These sensors, which include motion detectors, proximity sensors, and in some cases cameras, allow the system to continuously monitor the site or scarecrow facility. When the system detects intrusion or tampering, it triggers alert mechanisms such as alarms, notifications to farmers via mobile applications, and even automated deterrent measures. Integration of Arduino technology allows for a flexible and scalable system that can be adapted to a variety of agricultural environments. This study discusses the system's potential applications for different crop species and field sizes, highlighting its versatility and affordability for farmers with different needs and resources. Additionally, this project emphasizes the importance of real-time monitoring and proactive security measures in agricultural environments. The system aims to reduce crop damage caused by pests, wildlife or human intervention through intervention and early detection of potential threats, thereby optimizing crop protection and yield. That's what I'm aiming for. The use of Arduino-based technology has demonstrated the potential of using cost-effective and easily programmable microcontrollers in agricultural security systems. This not only improves crop protection, but also protects the scarecrow enclosure, which is essential to keep birds and pests out. The results of the research project are promising for farmers seeking efficient and affordable automated security solutions. It bridges the gap between traditional scarecrow mechanisms and modern technology, providing a practical and innovative approach to improving agricultural safety and productivity. Additionally, we highlight the potential for future development of smart agriculture by integrating Arduino-based systems to improve crop protection and scarecrow safety.

In [11] C.Szentpeteri's focus on the techniques to Tackle Bird Problems in the agricultural field Lasers and Drones Break Their Beaks, is a comprehensive examination of advanced technological solutions for bird control and highlights the effectiveness of lasers and drones in mitigating The focus of this study is on the introduction and analysis of innovative techniques specifically developed to solve bird-related problems. The authors examine the effectiveness and practicality of using lasers and drones as modern bird control tools in a variety of environments, including agricultural, industrial, and urban environments. Lasers are emerging as a promising method for bird deterrent. This study examines how a laser-based system can create visual disturbances and effectively disperse birds without causing physical harm. The precision and adaptability of lasers when targeting specific areas, such as grain fields or industrial sites, have been highlighted as key advantages over traditional methods. Additionally, this study also investigates the integration of drones into bird control strategies. Drones with specific features such as audiovisual deterrence and surveillance capabilities are being discussed as valuable tools for managing bird populations. Drones' ability to cover large areas, perform targeted interventions, and collect data for informed decision-making is highlighted, providing a dynamic and versatile approach to bird management. This study evaluates the benefits and limitations of these technologies. Lasers and drones have been recognized for their effectiveness in scaring away birds and preventing damage to crops, infrastructure, and other assets. However, it also addresses concerns such as operational costs, regulatory considerations, and the need for skilled operators, and highlights the practical challenges associated with widespread adoption. Furthermore, this study highlights the importance of an integrated approach to bird control and suggests that the most effective results can be achieved by combining technology and conventional methods. To optimize results, we advocate a comprehensive strategy that takes into account the specific situation, environmental factors, and behavior of the target bird species. Overall, this research project highlights the potential of lasers and drones as modern tools for bird conservation measures. It highlights the role of birds in providing more humane, efficient and targeted solutions to alleviate conflicts between birds and human activities in different regions. This research serves as a valuable resource for industry, agriculture, and urban planners seeking innovative, technology-driven approaches to addressing bird-related challenges.

[13] This approach is based on DNN. It consists of seven layers, two of which are connected and the other is a convolutional layer. Layers use Relu for nonlinear transformations. For max pooling, convolutional layers play an important role. The last layer is a regression layer that produces a binary object mask where the output has a centralized size. The value of the layer can be 1 and 0 where 1 means the pixel is within the bounding box of the layer, otherwise 0 is returned.

[12] The presented strategy describes a special approach for using convolutional neural networks (CNN) to generate classifiers with reduced robustness. Through the combination of these models, a more robust and diverse model will be created. Multi-view CNN (MV-CNN) technique is leveraged to process visual data efficiently by exploiting multiple projection views of the 3D model. Adjustments were made to the neural network architecture to support voxel data, with the training process covering multiple different directions for each element of the training dataset.
This approach makes it easy to capture features beyond those from pixel-based data.

.

The design of the Volumetric CNN envelops three convolutional layers working in three measurements, complemented by two completely associated layers, with the extreme layer serving as the classification module. The convolutional parts inside these layers are deliberately planned to capture voxel connections all through the profundity of the protest. By preparing the organize over various introductions for all models, an broad comprehension of spatial interconnects inside the protest is accomplished, traversing all tomahawks. To moderate computational costs, inadequate locally associated parts are utilized, capitalising on the three-dimensional nature of voxels to overcome spatial adjustment imperatives forced by customary two-dimensional CNNs due to part estimate confinements. Taking after the convolutional layer, the relu layer presents non-linearity, a basic component for compelling course separation. Along these lines, a pooling layer limits information excess, in this manner diminishing the model's estimate. A inclination for 3x3 parts is made within the Volumetric Convolutional Neural Arrange, as they demonstrate adequate for observing affiliations within voxelized information, apparent within the unmistakable division of isolated districts measuring 30 x 30 units. Dropout layers are judiciously incorporated to moderate overfitting dangers. Within the classification system, a completely associated layer with forty neurons is utilized for the ModelNet40 dataset, whereas a totally associated layer with ten neurons serves as the classifier for the ModelNet10 dataset. The FusionNet strategy coordinates different systems at the ultimate completely associated layer, yielding improved classification precision when compared to person arrange components.

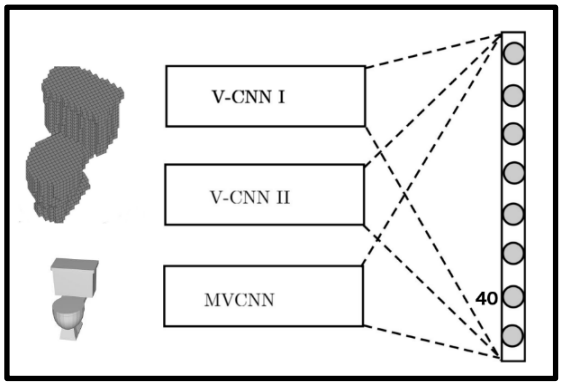

Fig 1: Two neural networks combination

The optimality isn't guaranteed as the base-network is non-convex. Assist the loss work must be regularized by adding and adjusting the weights on each yield that's generated. The items that are exceptionally little compared to the image size may well be relegated zero esteem to the yield that's gotten. To amend this the weights are adjusted and expanded for the comparison of non-zero values by parameter lambda. On the off chance that lambda is little, the values are punished when they are less than 1, which makes the design to estimate non-zero values indeed when the awful signals are experienced.

[15] The strategy is established in Deep Neural Systems (DNN), comprising an add up of seven layers, of which two are interconnected whereas the leftover portion are convolutional layers. Within the course of nonlinear change, each layer requires the use of Corrected Straight Units (ReLU). Convolutional layers play a significant part within the max-pooling method. The concluding layer constitutes a relapse layer that produces an Object-binary-mask with a centered dimensional yield. Outstandingly, the course esteem accept one of two states:

1, implying that the pixel dwells inside the class's bounding box, or 0, showing its area exterior said boundary. Given the non-convex nature of the foundational arrangement, accomplishing optimality remains questionable. Additionally, moderating the loss function's abnormality orders the alteration and alteration of weights connected to each generated output. To address this challenge, values essentially littler than the image size may be alloted a esteem of zero. In rectification, the weights experience variation and expansion for non-zero values through the presentation of a parameter signified as lambda. In occasions where lambda expects a little esteem, punishments are forced upon values less than one, actuating the engineering to expect values other than zero indeed within the nearness of weak signals.

[17] We created a coordinated neural arrangement for question discovery by combining all of the distinctive components into one. Our approach predicts person bounding boxes whereas moreover foreseeing bounding boxes for each category shows the interior of the picture utilizing image-wide attributes. This strategy permits the neural framework to do an in-depth evaluation of the complete picture and all its constituent objects. The YOLO engineering permits for end-to-end preparation whereas keeping up uncommon exactness and drawing nearer real-time preparing rates. Our strategy does this by isolating the input picture into a SS lattice. In case the midpoint of a thing falls inside a given bounded cell, it oversees identifying that question. Expectations for the boundary boxes of B and their individual network cells are created in each lattice cell.

**Implementation**

The process begins by generating a set of images of interest through an annotation process. Each creature is represented by approximately 50 images, and these visual representations are labeled accordingly. This labeling phase is important for instant detection because it assigns a unique label to each manual report. An application of transfer learning is achieved using the SSD-MobileNet model. The following steps describe the steps performed on the model.

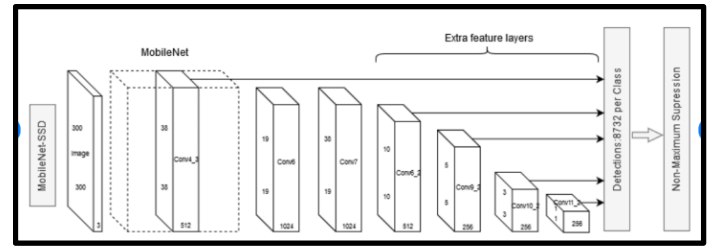

Fig 2 : MobileNet Architecture

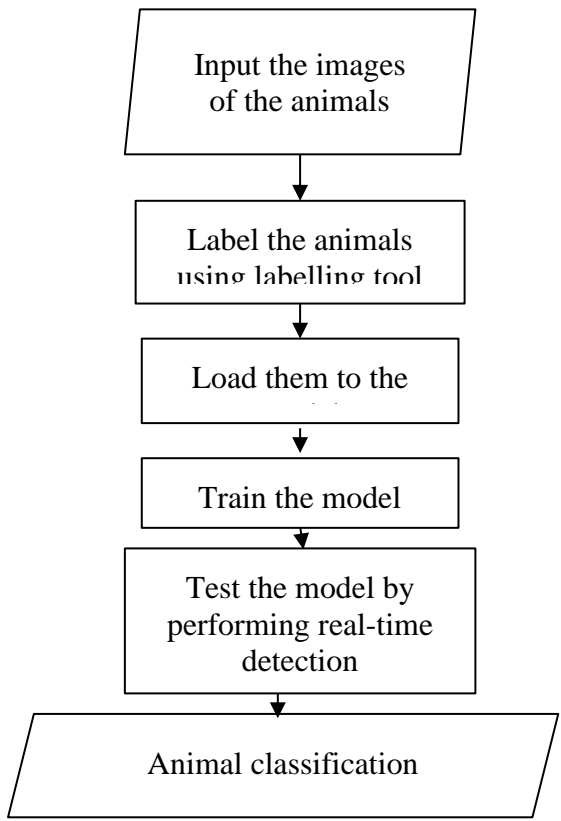

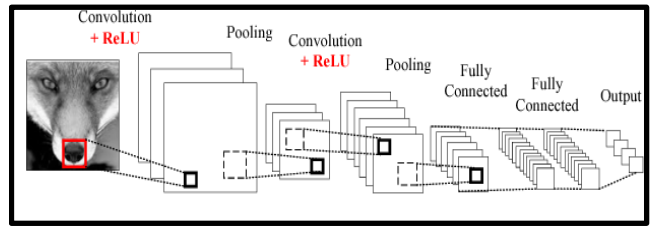

Fig 4 : Neural Network Architecture

Transfer learning is a machine learning approach that evaluates a system on large datasets toward a specific goal and then reuses or modifies it for another task. By implementing transfer learning in conjunction with the SSD MobileNet architecture, pre-built models are used for image classification and object identification. MobileNet stands out as a full-fledged deep learning framework for image classification, while SSD (Single Shot Detector) is an object identification framework. The next step will be instant detection using a webcam and his OpenCV library for object identification. Once image training and annotation are complete, the model is ready for recognition. The system is expected to achieve higher accuracy in identifying and classifying manual instructions, thereby providing reliable results.

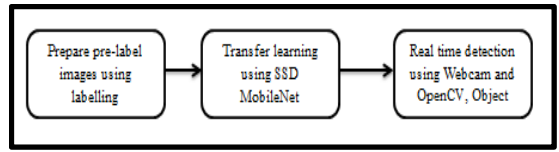

Fig 5: Working of model

The model is created by entering manual hints that may reveal symptoms of the specified condition. Manual entries are taken from both positive and negative classes. The details in these manuals are carefully labeled using the LabelImg tool. After the labeling process, this dataset is integrated into the model and subjected to a training program.

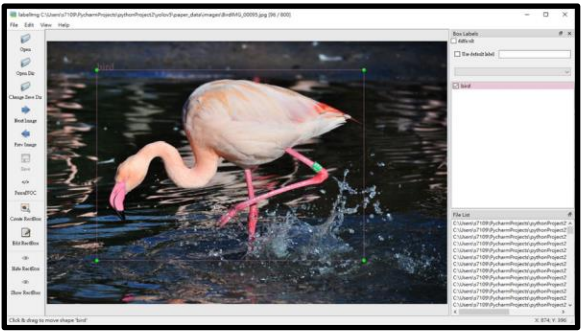

Fig 3: Labeling images using LabelImg

This training includes fine-tuning and optimization to improve model accuracy. In the final stage, the model undergoes rigorous real-time testing to assess its ability to distinguish between living things.

[19]The integration of advanced machine learning techniques into the domain of computer vision, exemplified by question detection within the TensorFlow Model Zoo, stands as a pivotal and highly lucrative innovation. This paradigm in technological advancement facilitates nuanced insights and precise localization of objects within both images and videos, offering transformative capabilities. The TensorFlow Model Zoo, boasting a repository of meticulously pre-trained models tailored explicitly for object detection, emerges as a practical and easily accessible solution across a diverse spectrum of real-world applications.The procedural initiation involves a meticulous selection process for a suitable pre-trained model from the TensorFlow library, with each model meticulously crafted to meet specific requirements, while thoughtfully balancing considerations such as precision and speed. Following this judicious selection, the designated model is seamlessly incorporated into the Python environment through TensorFlow's API, thereby facilitating the practical application of object detection to a designated dataset or project. These pre-trained models intrinsically possess the

capability to discern and categorize a diverse array of objects, rendering them indispensable for various applications, including autonomous driving, surveillance, medical imaging, and more. The TensorFlow Model Zoo effectively transforms object detection into a readily available and efficient tool, streamlining the assimilation of sophisticated computer vision solutions across various industries and use cases.

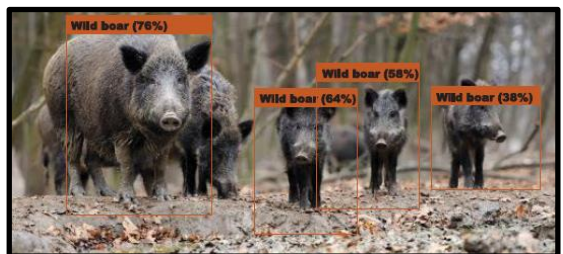

Fig 6 : Object Detection

In the realm of image processing, images undergoing the training phase undergo meticulous preprocessing to extract relevant information. These files serve as comprehensive repositories of the extracted characteristics, providing a benchmark for subsequent testing endeavors. Following preprocessing, features are systematically extracted using the previously trained model, and these features are systematically juxtaposed with the stored ones to ascertain the closest match or similarity with a recognized gesture. This systematic approach enables the real-time detection of the animals, wherein the identification of a gesture prompts the recognition.

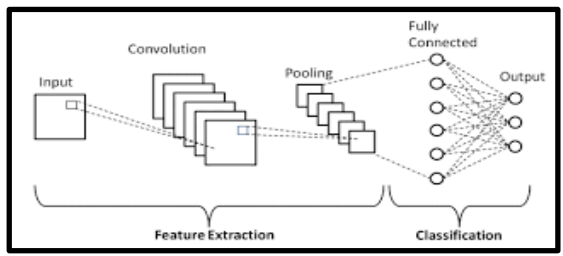

Fig 7 : Feature extraction and Classification

The method commences with the start of input gathering at the primary layer, wherein the starting introduction of the input goes before its transmission to the concealed layers. These darkened layers envelop a huge number of convolutional layers with differing sizes and amounts. Collectively, these convolutional layers contribute significantly to deciding an interesting certainty esteem. Within the assessment of the model's adequacy, a testing arrangement traversing 10,000 steps was built up, utilizing indistinguishable hyperparameters to those utilized amid the preparing stage. All through the preparing regimen, the demonstration encounters diminutions in categorization and regularization. The localization misfortune serves as a metric evaluating the contrast between the expected and genuine improvements in bounding boxes. This misfortune is unexpected upon adjustments made to the shown bounding boxes and their genuine values. Pivotal parameters incorporate the overall tally of predefined holders (N), anticipated constrain holders (l), genuine encasing holder (g), encoded representation of the genuine encasing holder (g^), and compatibility measurements between predefined and genuine encasing holders for category k (xij^k). A comprehensive talk with respect to the brought about misfortunes amid experimentation is elucidated in a consequent area, citing source.Ensuring the preparing stage, the framework is started, commencing from the most recent checkpoint, in this manner enabling real-time location. Configurational alterations and updates are executed to prepare the demonstration for directions arrangement. The prepared show is at that point stacked with the foremost later checkpoint, concluding the preliminary stages for real-time sign dialect acknowledgment. This preparation is versatile for advance customization to meet particular necessities, including considerations related to the integration of nearby dialect complexities.

## Results and discussions

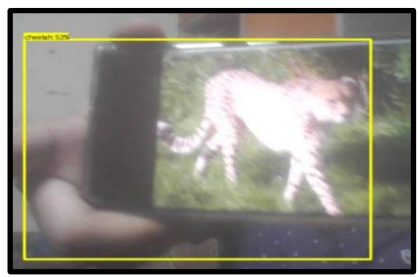

Fig 8 : Detection of Cheetah

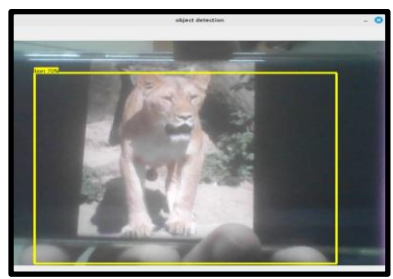

Fig 9 : Detection of Lion

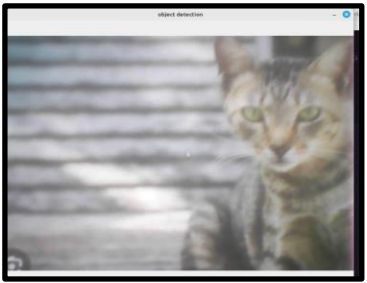
Fig 10 : Detection of cat

When implementing a scarecrow monitoring system using the MobileNet SSD model, it reflects the system's ability to accurately predict and distinguish wild animals. In particular, wild animals such as lions were accurately identified, consistent with the system's goal of monitoring potential threats to crops from larger and more dangerous animals. In contrast, domestic animals such as cats were not accurately identified because they are not considered wild and pose less of a threat to agricultural land. These accurate predictions ensure that the system effectively targets animals that can harm crops, demonstrating the practicality and reliability of the scarecrow monitoring system in real-world agricultural scenarios.

Agriculture is currently undergoing a technological revolution, and one of the breakthrough innovations is the Scarecrow Monitoring System (SMS). Computer vision integration. At the heart of this system is the use of the MobileNet SSD model, known for its lightweight architecture and real-time processing capabilities. This comprehensive overview delves into the complexity of the project and highlights key aspects such as the training phase, real-time detection, accuracy metrics, and system efficiency. It brings technology and agriculture together , the Scarecrow Surveillance System addresses the urgent need for advanced tools in agriculture to protect crops through the use of cutting-edge computer vision technology. The purpose is that. The focus of this project is the use of the MobileNet SSD model, chosen for its optimal balance between accuracy and computational efficiency. SMS has proven to be a promising solution for farmers looking to improve their crop protection measures. The success of any computer vision system depends on the effectiveness of the training phase. The project took a careful approach, with each species trained on his dataset of **30** images. The importance of this approach lies in the ability to deeply understand various visual properties within the MobileNet SSD model. The diversity of the training dataset, including different angles, lighting conditions, and backgrounds, was found to be critical to the model's ability to generalize effectively. The selection of the MobileNet SSD model further contributed to the success of the training phase. Its lightweight architecture facilitates efficient learning, allowing the model to recognize important features and patterns associated with each animal species. This phase resulted in a well-trained model that allows for accurate and reliable animal identification. The Scarecrow monitoring system is characterized by its real-time detection capability, a key feature for timely intervention in situations that may threaten crops. By leveraging the MobileNet SSD model, the system demonstrated excellent ability to accurately identify animals moving through the surveillance area. This real-time processing is invaluable to farmers, providing timely information to proactively respond and reduce the risk of crop damage. An accuracy of approximately **91%** was achieved, proving the effectiveness of his MobileNet SSD model in identifying different animal species. The model's ability to recognize key features acquired during the training phase resulted in a high level of accuracy when deployed in the real world. This accuracy not only makes the system reliable, but also makes it a valuable tool for farmers seeking advanced techniques for crop protection. The Scarecrow surveillance system is superior not only in terms of accuracy but also in terms of efficiency, which is an important indicator of the system's practicality. The lightweight architecture of MobileNet SSD models ensures optimal performance even on resource-constrained devices, contributing to system cost-effectiveness and scalability. This efficiency is critical for continuous field monitoring, providing farmers with a reliable and responsive animal monitoring tool. Efficiency metrics were evaluated in terms of both computing resources and processing speed. MobileNet's SSD models exhibit an excellent balance of accuracy and speed, making them ideal for real-time applications. The system's ability to maintain high accuracy when operating in real-time scenarios highlights its efficiency, an important characteristic for the continued success of a scarecrow surveillance system. In summary, the Scarecrow monitoring system using the MobileNet SSD model has proven to be a game changer in the field of precision agriculture. The combination of carefully selected training datasets, MobileNet SSD model implementation, and the system's real-time detection capabilities has resulted in a solution that meets the evolving needs of modern agriculture. This not only helps in the accurate identification of animals, but also lays the foundation for further research into advanced technologies in agriculture. As the interface between technology and agriculture continues to evolve, scarecrow monitoring systems are at the vanguard of innovation and offer a glimpse into the future of sustainable and efficient farming practices.

**Conclusion and Future Works**
The computer vision technology with traditional scarecrow systems provides real-time monitoring and precise intervention strategies to achieve optimization. Crop protection. The outstanding performance of this system is its amazing accuracy in detecting pests and intruders. Powered by MobileNet SSD, the system quickly and accurately detects threats in agricultural areas and significantly minimizes false alarms. This precision allows the scarecrow system to effectively respond to real risks and reduce potential crop damage and loss. Real-time monitoring is an important feature.The instant notification allows for rapid assessment and targeted response, ultimately increasing crop yields and reducing agricultural losses. Resource optimization represents an important advance in project success. MobileNet SSD's lightweight design maximizes computing efficiency and enables use in edge devices with limited processing power. This adaptability ensures scalability in a variety of agricultural environments, facilitating accessibility and widespread adoption in the farming community. Cost efficiency was found to be a significant advantage. MobileNet SSD's optimized neural network model minimizes hardware requirements and associated costs, making intelligent Scarecrow monitoring systems economically viable for farmers with various financial constraints. This affordability extends its reach and allows a wider range of farmers to take advantage of the system's benefits.Future works would be involved with an intuitive dashboard that displays monitoring results in real time, the system is aimed at users with a variety of technical skills. This ease of use enables informed decision-making and enables farmers to respond quickly and effectively to the insights provided by the scarecrow monitoring system.Its targeted interventions promote environmentally friendly agricultural practices, reducing environmental impact while protecting crops.

## References

[1] W. Dai, H. Nishi, V. Vyatkin, V. Huang, Y. Shi, and X. Guan, ''Industrial edge computing: Enabling embedded intelligence,'' IEEE Ind. Electron. Mag., vol. 13, no. 4, pp. 48–56, Dec. 2019.

[2] A. Roihan, M. Hasanudin, and E. Sunandar, ''Evaluation methods of bird repellent devices in optimizing crop production in agriculture,'' J. Phys., Conf. Ser., vol. 1477, Mar. 2020, Art. no. 032012.

[3] N. P. Ramaiah, E. P. Ijjina, and C. K. Mohan, ''Illumination invariant face recognition using convolutional neural networks,'' in Proc. IEEE Int. Conf. Signal Process., Informat., Commun. Energy Syst. (SPICES), Feb. 2015, pp. 1–4.

[4] Aishwarya, K.; Kathryn, J.C.; Lakshmi, R.B. A Survey on Bird Activity Monitoring and Collision Avoidance Techniques in Windmill Turbines. In Proceedings of the 2016 IEEE Technological Innovations in ICT for Agriculture and Rural Development (TIAR), Chennai, India, 15–16 July 2016; pp. 188–193.

[5] M. S. Farooq, S. Riaz, A. Abid, K. Abid, and M. A. Naeem, ``A survey on the role of IoT in agriculture for the implementation of smart farming,'' IEEE Access, vol. 7, pp. 156237- 156271, 2019.

[6] S. Giordano, I. Seitanidis, M. Ojo, D. Adami, and F. Vignoli, ``IoT solutions for crop protection against wild animal attacks,'' in Proc. IEEE Int. Conf. Environ. Eng. (EE), Mar. 2018, pp.

[7] Prabhakar, T., Srujan Raju, K., Reddy Madhavi, K. (2022). Support Vector Machine Classification of Remote Sensing Images with the Waveletbased Statistical Features. Fifth International Conference on Smart Computing and Informatics (SCI 2021), Smart Intelligent Computing and Applications, Volume 2. Smart Innovation, Systems and Technologies, vol 283. Springer, Singapore.

[8] S. Yadahalli, A. Parmar and A. Deshpande, "Smart Intrusion Detection System for Crop Protection by using Arduino," 2020 Second International Conference on Inventive Research in Computing Applications (ICIRCA), Coimbatore, India, 2020, pp. 405-408, DOI:10.1109/ICIRCA48905.2020.9182868.

[9] ] D. Shadrin, A. Menshchikov, D. Ermilov, and A. Somov, ''Designing future precision agriculture: Detection of seeds germination using artificial intelligence on a low-power embedded system,'' IEEE Sensors J., vol. 19, no. 23, pp. 11573–11582, Aug. 2019.

[10] G. Codeluppi, L. Davoli, and G. Ferrari, ''Forecasting air temperature on edge devices with embedded AI,'' Sensors, vol. 21, no. 12, p. 3973, Jun. 2021.

[11] Szentpeteri, C. Bird Control: Technology to Tackle Your Bird Troubles: Lasers and Drones Beat the Beak. Aust. N. Z. Grapegrow. Winemak. 2018, 657, 31.

[12] Yoshihashi, R.; Kawakami, R.; Iida, M.; Naemura, T. Bird Detection and Species Classification with Time-Lapse Images around a Wind Farm: Dataset Construction and Evaluation. 2017.

[13] C Szegedy, Alexander Toshev and Dumitru Erhan, "Deep Neural Networks for object detection", (January 2013)

[14] Zeiler, Matthew D, and Rob Fergus.. "Visualizing and Understanding Convolutional Networks." (2014)

[15] Sermanet, Pierre "Overfeat: Integrated Recognition, Localization and Detection Using Convolutional Networks." (2013)

[16]Shih, Ya-Fang "Deep Co-Occurrence Feature Learning for Visual Object Recognition." In Proc. Conf. Computer Vision and
Pattern Recognition.(2017)

[17] Redmon, Joseph, Santosh Divvala, Ross Girshick, and Ali Farhadi,"You Only Look Once: Unified, Real-Time Object Detection." (2016)

[18] Elisabeth Brogren , Lars B. Dahlin  Erika Franzen, "Striatal Hand Deformities in Parkinson's Disease: Hand Surgical Perspectives" (2022)

[19] "Tensorflow Zoo model for object detection", Towards Data Science site,(2022)